\documentclass[10pt,twocolumn,letterpaper]{article}

\usepackage{iccv}
\usepackage{times}
\usepackage{epsfig}
\usepackage{graphicx}
\usepackage{amsmath}
\usepackage{amssymb}

\usepackage[utf8]{inputenc}
\usepackage{xcolor}
\usepackage{mathtools}
\usepackage{bm}
\newcommand{\norm}[1]{\left\lVert#1\right\rVert}

\usepackage[pagebackref=true,breaklinks=true,letterpaper=true,colorlinks,bookmarks=false]{hyperref}

\iccvfinalcopy 

\def\iccvPaperID{7515} 

\ificcvfinal\fi

\begin{document}

\title{Long-Term Temporally Consistent Unpaired Video Translation from Simulated Surgical 3D Data}

\author{
Dominik Rivoir\textsuperscript{1,2}, Micha Pfeiffer\textsuperscript{1}, Reuben Docea\textsuperscript{1}, Fiona Kolbinger\textsuperscript{1,3},\\
Carina Riediger\textsuperscript{3}, Jürgen Weitz\textsuperscript{2,3}, Stefanie Speidel\textsuperscript{1,2}
\and
\textsuperscript{1}NCT/UCC Dresden, Germany,
\textsuperscript{2}CeTI, TU Dresden,
\textsuperscript{3}University Hospital Dresden\\
{\tt\small \{dominik.rivoir, micha.pfeiffer, reuben.docea, stefanie.speidel\}@nct-dresden.de}\\
{\tt\small \{fiona.kolbinger, carina.riediger, juergen.weitz\}@uniklinikum-dresden.de}
}

\maketitle
\ificcvfinal\thispagestyle{empty}\fi

\begin{abstract}
Research in unpaired video translation has mainly focused on short-term temporal consistency by conditioning on neighboring frames.
However for transfer from simulated to photorealistic sequences, available information on the underlying geometry offers potential for achieving global consistency across views.
We propose a novel approach which combines unpaired image translation with neural rendering to transfer simulated to photorealistic surgical abdominal scenes.
By introducing global learnable textures and a lighting-invariant view-consistency loss, our method produces consistent translations of arbitrary views and thus enables long-term consistent video synthesis.
We design and test our model to generate video sequences from minimally-invasive surgical abdominal scenes. Because labeled data is often limited in this domain, photorealistic data where ground truth information from the simulated domain is preserved is especially relevant. By extending existing image-based methods to view-consistent videos, we aim to impact the applicability of simulated training and evaluation environments for surgical applications. Code and data:
\url{http://opencas.dkfz.de/video-sim2real}.
\end{abstract}

\begin{figure}[t]
\begin{center}
\includegraphics[width=.9\linewidth, trim=10.1cm 1.2cm 9.5cm .3cm, clip]{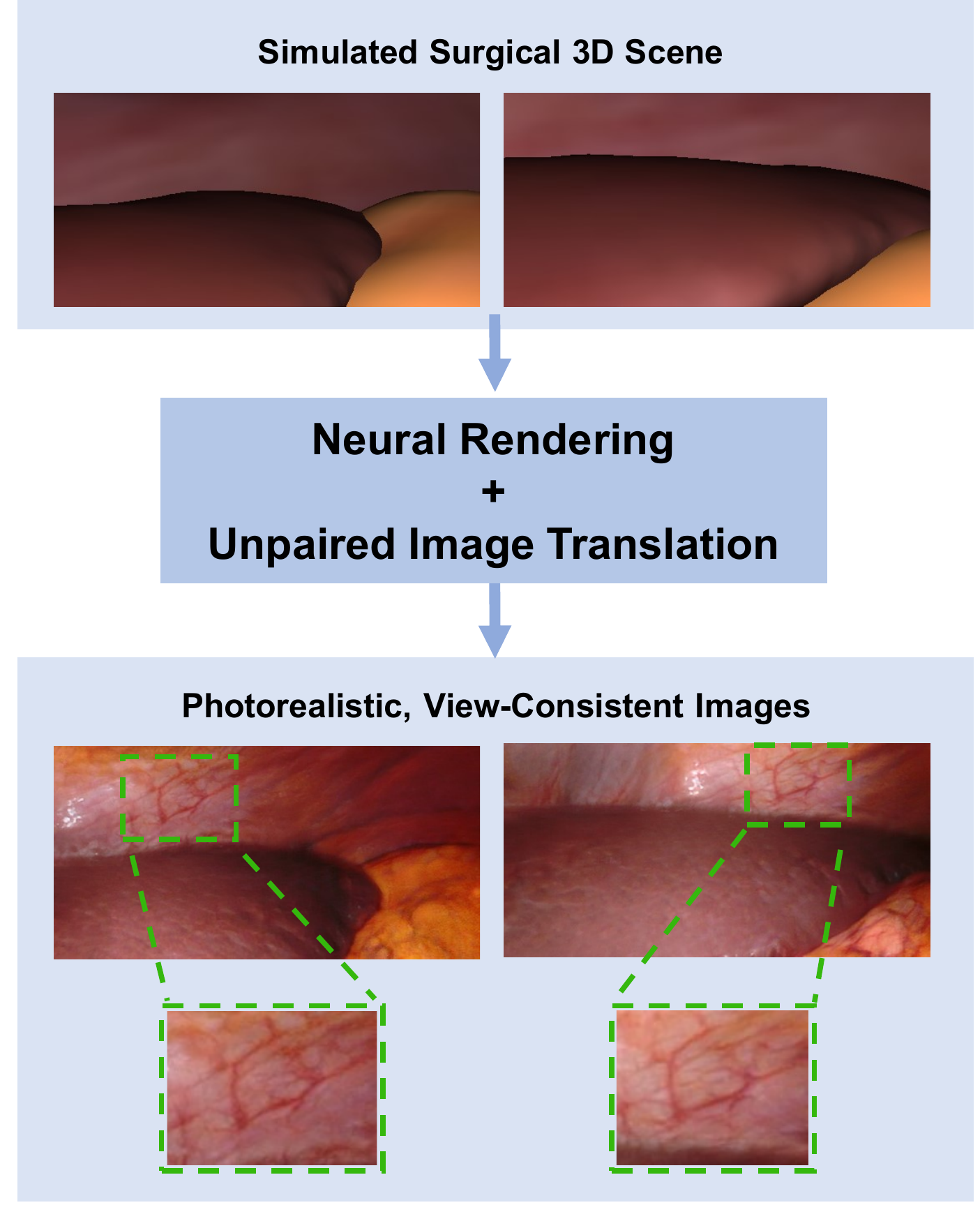}
\end{center}
   \caption{By combining unpaired image translation with a neural rendering approach, we produce photorealistic and view-consistent renderings of simulated surgical scenes. Note that fine details like vessels are rendered consistently across viewpoints although they were not manually modelled in the simulated domain.}
\label{fig:concept}
\end{figure}

\section{Introduction}
One of the most promising applications of GAN-based image translation~\cite{goodfellow2014generative,wang2018high} is the transfer from the simulated domain to realistic images as it presents great potential for applications in computer graphics. More importantly, unpaired translation~\cite{zhu2017unpaired} (\ie no image correspondences between domains required during training) enables the generation of realistic data while preserving ground information from the simulated domain which would otherwise be difficult to obtain (\eg depth maps, optical flow or semantic segmentation). This synthetic data can then facilitate training or evaluation in settings where labeled data is limited.

The availability of realistic, synthetic data is especially crucial in the field of computer-assisted surgery (CAS)~\cite{maier2017surgical,bodenstedt2020artificial}. CAS aims at providing assistance to the surgical team (\eg visualizing target structures or prediction of complications) by means of analyzing available sensor data. In minimally-invasive surgery, where instruments and a camera are inserted into the patient's body through small ports, video is the predominant data source. Intelligent assistance systems are especially relevant here, since performing surgery through small ports and limited view is extremely challenging. However, two major factors which currently limit the impact of deep learning in CAS are the lack of (a) labeled training data and (b) realistic environments for evaluation~\cite{maier2020surgical}. For instance, evaluating a SLAM (Simultaneous Localization and Mapping) algorithm~\cite{mur2015orb,song2018mis} on laparoscopic video data poses several problems since the patient's ground truth geometry is typically not accessible in the operating room (OR) and recreating artificial testing environments with realistic and diverse patient phantoms is extremely challenging. Other CAS applications which could benefit from temporally consistent synthetic training data include action recognition, warning systems, surgical navigation and robot-assisted interventions~\cite{maier2017surgical,maier2020surgical}.

Previous research has shown the effectiveness of synthetic, surgical images as training data for downstream tasks such as liver segmentation~\cite{pfeiffer2019generating,sahu2020endo}. However, their applications are still limited since many challenges in CAS include a temporal component. Using the previous example of evaluating a SLAM algorithm, realistic as well as temporally consistent video sequences would have to be generated in order to provide a useful evaluation environment.

Unpaired video translation has recently garnered interest in various non-surgical specialties~\cite{bansal2018recycle,engelhardt2018improving,chen2019mocycle,chu2020learning,park2019preserving,xu2020ofgan}. Most approaches thereby condition the generator on previous translated frames to achieve smooth transitions, \ie short-term temporal consistency. However, they are fundamentally not designed for long-term consistency. Intuitively, when an object entirely leaves the field of view, consistent rendering cannot be ensured when it returns since the previous frame contains no information regarding the object's appearance. Even when the model is conditioned on multiple frames, the problem persists in longer sequences.

In the special case of translating from a simulated environment, however, the underlying geometry and camera trajectories are often available. Point correspondence between views are thus known and can be used to ensure globally consistent translations. The relatively new research area of neural rendering~\cite{tewari2020state} aims at using the knowledge of the underlying 3D scene for image synthesis but has mainly been studied in supervised settings to date~\cite{tewari2020state,lombardi2019neural,sitzmann2019deepvoxels,thies2019deferred,mildenhall2020nerf}.

We propose a novel approach for unpaired video translation which utilizes the available information of the simulated domain's geometry to achieve long-term temporal consistency. A state-of-the-art image translation model is extended with a neural renderer which learns global texture representations. This way, information can be stored in 3D texture space and can be used by the translation module from different viewpoints. \emph{I.e.} the model can learn the position of details such as vessels and render them consistently (Fig. \ref{fig:concept}). To ensure texture consistency, we introduce a lighting-invariant view-consistency loss. Furthermore, we employ methods to ensure that labels created in the simulated domain remain consistent when translating them to realistic images.
We show experimentally that our final generated video sequences retain detailed visual features over long time distances and preserve label consistency as well as optical flow between frames.

\section{Related Work}

\subsection{Unpaired Image and Video Translation}
Image-based GANs~\cite{goodfellow2014generative,radford2015unsupervised} have gathered much attention showing impressive results as unconditioned generative models~\cite{radford2015unsupervised,brock2018large,karras2017progressive,karras2019style,karras2020analyzing} or in conditional settings such as image-to-image translation~\cite{wang2018high,choi2018stargan,park2019semantic,liu2019learning}. However, their real-world applications are limited, since the content of generative models is difficult to control and supervised image translation requires corresponding image pairs, which are often not available. The introduction of unpaired translation through cycle consistency~\cite{zhu2017unpaired} hence widened their applicability and impact. Since then, several extensions have been proposed, \eg shared content spaces~\cite{liu2017unsupervised}, multi-modality~\cite{lee2018diverse,huang2018multimodal}, few-shot translation~\cite{liu2019few} or replacing cycle consistency with contrastive learning~\cite{park2020contrastive}. From an application standpoint, several works~\cite{pfeiffer2019generating,mathew2020augmenting,sahu2020endo,rau2019implicit,marzullo2021towards,widya2021self} have shown the effectiveness of leveraging synthetic training data for surgical applications.

There have been several attempts at extending unpaired translation to videos where generated sequences have to be temporally smooth in addition to being realistic in individual frames~\cite{bansal2018recycle,engelhardt2018improving,chen2019mocycle,chu2020learning,park2019preserving,xu2020ofgan}. Bansal \etal~\cite{bansal2018recycle} tackle this problem by introducing a temporal cycle consistency loss and Engelhardt \etal~\cite{engelhardt2018improving} use a temporal discriminator to model realistic transitions between frames. Several recent approaches estimate optical flow to ensure temporal consistency in consecutive frames~\cite{chen2019mocycle,chu2020learning,park2019preserving,xu2020ofgan}. While there have been steady improvements in generating smooth transitions between frames, these models fail to capture long-term consistency. We aim to overcome this by adding a neural rendering component to our model. To our best knowledge, no successful solutions for long-term consistent video translation in the unpaired setting have been published to date.

\begin{figure*}
\begin{center}
\includegraphics[width=\linewidth, trim= 1cm 2.9cm .7cm 2.3cm, clip]{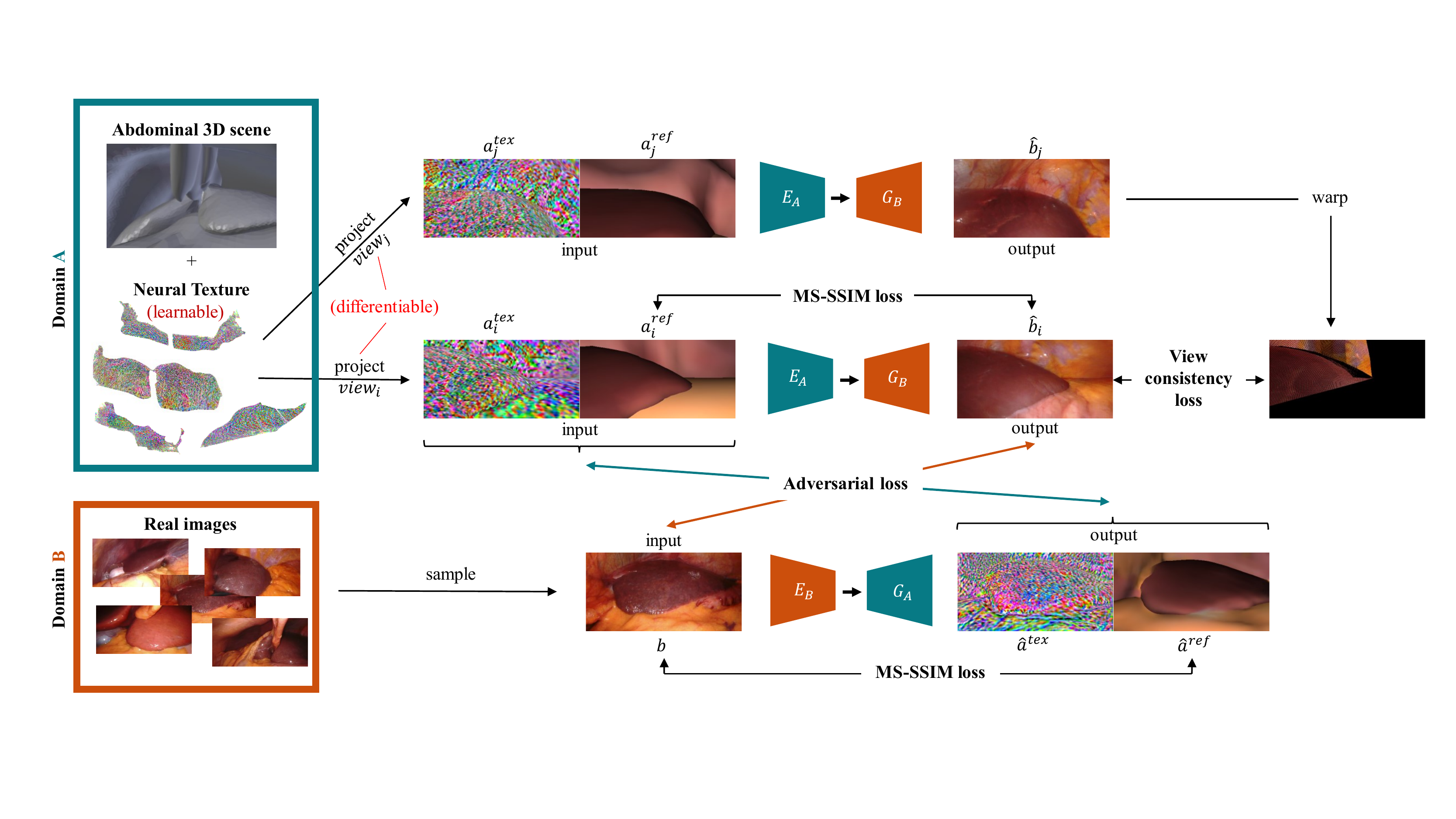}
\end{center}
\caption{We combine unpaired image translation with neural rendering for view-consistent translation from simulated to photorealistic surgical videos. The model's key concept is a learnable, implicit representation of the scene's global texture. During training, texture features are projected into image space as $a^{tex}_i$ which, combined with a simple rendering $a^{ref}_i$, serve as input to the unpaired image translation module. To encourage long-term temporally consistent translation, we warp two translated views into a common pixel space and employ our lighting-invariant consistency loss. Also note that the projected texture maps are part of the translation cycle, \ie transfer from $B$ to $A$ includes the prediction of a reference image $\hat{a}^{ref}$ as well as a texture map $\hat{a}^{tex}$.}
\label{fig:model}
\end{figure*}

\subsection{Physically-grounded Neural Rendering}
While unpaired visual translation methods are also sometimes categorized as neural rendering, the term most commonly refers to image synthesis approaches which incorporate knowledge of the underlying physical world~\cite{tewari2020state}. By introducing differentiable components to rendering pipelines, neural representations of 3D shapes~\cite{mildenhall2020nerf,sitzmann2019deepvoxels,zhu2018visual,lombardi2019neural}, lighting~\cite{sun2019single,meka2019deep,alhaija2018geometric,alhaija2020intrinsic}, textures~\cite{thies2019deferred} or view-dependent appearance~\cite{mildenhall2020nerf} can be learned from image data for applications like novel view synthesis, facial re-enactment or relighting. Most closely related to our work, Thies \etal~\cite{thies2019deferred} introduce a \emph{deferred neural renderer} with \emph{neural textures}, where implicit texture representations are learned from image sequences with a ground truth 3D model and camera poses. In contrast to their work, however, our model is built in an unsupervised setting where no correspondence between the simulated 3D data and real images is available. Finally, Alhaija \etal~\cite{alhaija2020intrinsic} propose a \emph{deferred neural renderer} for unpaired translation from fixed albedo, normal and reflection maps to realistic output. However, since the texture representation is not learned, this work is more closely related to image-to-image translation.

Mallya \etal~\cite{mallya2020world} propose a model for long-term consistent, paired video translation. They estimate information of the underlying physical world (depth, optical flow, semantic segmentation) to render globally consistent videos. This is currently the closest attempt at combining neural rendering with GAN-based translation. However, the paired data setting is a major hurdle for real-world applications.

We rather aim at bringing unpaired translation and neural rendering closer together. We believe that requiring knowledge of the simulated 3D geometry in an unpaired setting is often less restrictive than paired video translation, which requires rich ground truth in the real domain.

\section{Method}
We propose a method for unpaired, view-consistent translation from the domain of simulated surgical scenes $A$ to the domain of realistic surgical images $B$. The method consists of three components: learnable neural textures, an unpaired image translation module and a lighting-invariant view-consistency loss (Fig.~\ref{fig:model}).

$1)$ For a given viewpoint $view_i$ of the simulated scene, learnable features are projected from the neural texture $tex$ to their pixel locations in the image plane and form a spatial feature map $a^{tex}_i$:
\begin{equation} \label{eq:project}
    a^{tex}_i = project(tex,view_i)
\end{equation}

$2)$ Additionally, a simple but unrealistic rendering $a^{ref}_i$ of the same view is used as a prior for translation.
Combined, $(a_i^{tex},a_i^{ref}) \in A$ serves as input to the unpaired image translation module to get the fake image $\hat{b}_i \in B$:
\begin{equation}
    \hat{b}_i = translate_\theta(a^{ref}_i,a^{tex}_i)
\end{equation}
Errors can be backpropagated into $tex$ since $project(\cdot)$ is differentiable and enables the model to learn the global texture representation $tex$ end to end with the network parameters $\theta$ of the translation module.

$3)$ To ensure globally consistent rendering, pairs of translated views $\hat{b}_i,\hat{b}_j$ are sampled during training, warped into a common pixel space and constrained using our lighting-invariant view-consistency loss.
\begin{equation}
    \hat{b}_i \xleftrightarrow[consist.]{view} \hat{b}_j
\end{equation}

The main insight of our model is that neural textures allow the model to learn global information about the scene independent of time-point and view, \eg material properties or locations of details such as vessels. After projecting texture features into the image plane, the translation module serves as a deferred renderer to synthesizing realistic images. Since the translation module operates on individual views, view-dependent effects such as specular reflections or changing lighting conditions are synthesized here.
We jointly learn one neural texture for each of the 7 simulated scenes and common translation networks for all scenes.

\subsection{Neural Texture \& Projection Mechanism}
For true long-term consistency, we require a method which can store information independent of time-point or view. To do so, we use a learnable, global texture $tex$, named \emph{neural texture} by Thies et al.~\cite{thies2019deferred}. For each object in the scene (liver, gallbladder, ligament, abdominal wall and fat), $tex$ contains learnable, spatial feature maps as an implicit texture representation. At each spatial location (texel), $N$ features are learned and enable the model to learn consistent tissue properties or locations of details such as vessels. The shape of $tex$ is $O \times P \times H \times W \times N$ with $O=5$ objects, $P=6$ projection planes of size $H\times W = 512\times 512$ per object and $N=3$ learnable texture features per texel.

To learn $tex$ end-to-end with the translation module, we only require a differentiable projection mechanism (Eq.~\ref{eq:project}) which maps features from the global texture $tex$ into the image plane for a given view $view_i$. The resulting image-sized feature map $a^{tex}_i$ serves as an input to the translation module and thus errors can be propagated back into $tex$.
\begin{equation} \label{eq:triplanar}
    a_i^{tex}[x,y] = \sum_{x_p,y_p}^{tri(s)} (n_s^T \cdot n_p)^2 \cdot tex[o,p,x_p,y_p]
\end{equation}

\begin{figure}[t]
\begin{center}
    \includegraphics[width=\linewidth, trim= 0cm 1.3cm 0cm .8cm, clip]{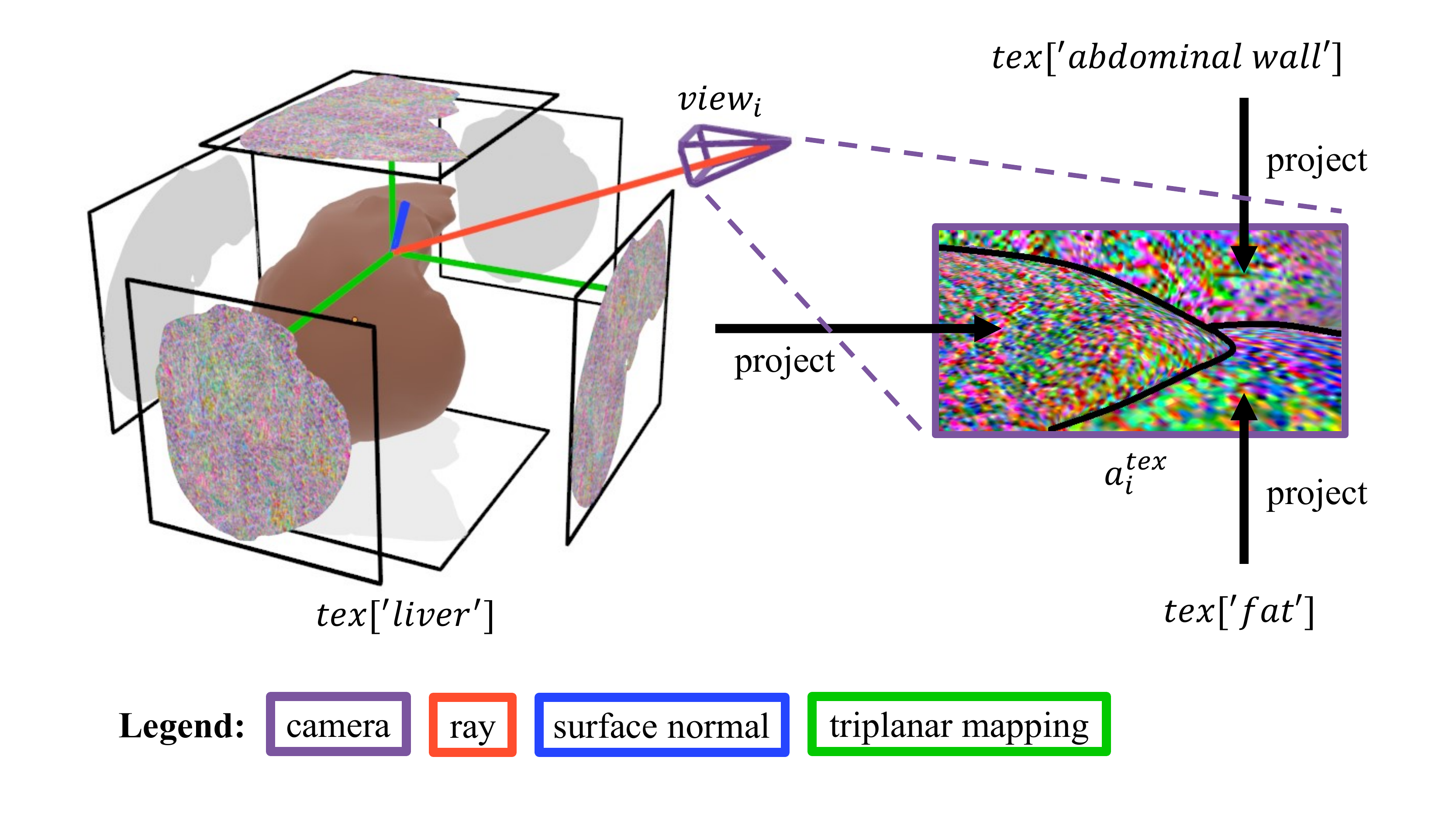}
\end{center}
   \caption{Neural texture projection. A ray (red) is cast for every pixel to the scenes's surface. Triplanar mapping (green) is used to map surface points to learnable texture planes (with bilinear interpolation in texture space). This differentiable mapping allows us to backpropagate errors from image space to global texture space.}
\label{fig:triplanar}
\end{figure}

We define the projection into $a^{tex}_i$ by means of ray casting~\cite{watt1992advanced}, triplanar mapping~\cite{golus2017normal} and bilinear interpolation~\cite{gibson2000handbook} (Fig. \ref{fig:triplanar} and Eq.~\ref{eq:triplanar}).
For each pixel $(x,y)$, we cast a \emph{ray} onto its $3D$ surface point $s\in\mathbb{R}^3$ in the scene and determine the object $o$ it belongs to. The neural texture $tex[o]$ of an object consists of 6 axis-aligned texture planes surrounding the mesh.
Through \emph{triplanar mapping} $tri(s)$, we obtain one texture coordinate $(x_p,y_p)$ for each of the three planes $p \in \{1..P\}$ which face $s$ (Fig.~\ref{fig:triplanar}).
Texture features are weighted by the dot-product of plane and surface normal $n_p,n_s$ to obtain the aggregated features $a_i^{tex}[x,y]$ in pixel space. Since texture planes are discrete grids, we use \emph{bilinear interpolation} to obtain texture features $tex[o,p,x_p,y_p]$ from arbitrary, continuous locations. Hence, a total of $12$ texels contribute to one pixel ($4$ discrete texels for each of the 3 plane coordinates). For details, see the supplementary.
Note that triplanar mapping was chosen for its simplicity but could easily be replaced by other UV mappings.

\subsection{Unpaired Image Translation Module}
Our translation module is a deterministic, style-less variant of Pfeiffer et al.'s model~\cite{pfeiffer2019generating}, which itself is based on MUNIT~\cite{huang2018multimodal}. The model enforces cycle consistency as well as a shared content space through interchangeable encoders $E_A,E_B$ and decoders $G_A,G_B$ for each domain~\cite{liu2017unsupervised}.

Given a projected texture map and reference image $(a^{tex}_i,a^{ref}_i) \in A$, the encoder $E_A$ extracts a domain-independent content code $c^{a_i}$ and decoder $G_B$ predicts a fake image $\hat{b}_i \in B$ from $c^{a_i}$. $E_B$ then reconstructs content $c^{a_i}_{rec}$ and $G_A$ translates back to domain $A$ to complete the cycle. Additionally, the input is directly reconstructed through $(a_{i,rec}^{tex},a_{i,rec}^{ref})=G_A(E_A(a^{tex}_i,a^{ref}_i))$. Translation from $B$ to $A$ is done analogously. Finally, Multi-Scale Discriminators~\cite{wang2018high} $D_A, D_B$ distinguish fake and real images.
\begin{equation}
    \begin{pmatrix}
    a^{tex}_i\\
    a^{ref}_i
    \end{pmatrix}
    \xrightarrow{E_A}
    c^{a_i}
    \xrightarrow{G_B}
    \hat{b}_i
    \xrightarrow{E_B}
    c^{a_i}_{rec}
    \xrightarrow{G_A} 
    \begin{pmatrix}
    a_{i,cyc}^{tex}\\
    a_{i,cyc}^{ref}
    \end{pmatrix}
\end{equation}
We use the LS-GAN loss~\cite{mao2017least} as adversarial loss $L_{adv}$, and $L1$ losses for $L_{cyc}, L_{rec}, L_{c}$ to ensure cycle consistency as well as image and content reconstruction.
Finally, we enforce a Multi-Scale Structural-Similarity loss~\cite{wang2003multiscale,pfeiffer2019generating} $L_{ssim}$ on the brightness of $a_i^{ref}$ and $\hat{b}_i$ as well as $b$ and $\hat{a}^{ref}$ to encourage label-preserving translation. Details on networks and losses can be found in the supplementary.
\begin{equation}
    L_{translation} = L_{adv} + L_{cyc} + L_{rec} + L_{c} + L_{ssim}
\end{equation}


\subsection{View Consistency Loss}
To enforce view consistency, two random views $i,j$ of the same simulated scene are sampled and translated during each training iteration. Using the knowledge about the scene's geometry, the second view is warped into the pixel space of the first view and consistent rendering is enforced through a pixel-wise view-consistency loss. In minimally-invasive surgery, however, the only source of light is a lamp mounted on the camera. This results in changing light conditions whenever the field of view is adjusted and the image center typically being brighter than its surroundings. This poses an additional challenge for view-consistency. Therefore, we propose to minimize the angle between RGB vectors instead of a channel-wise loss. For a pair of translated views $\hat{b}_i, \hat{b}_j$, the loss is defined as
\begin{equation}
L_{vc} = \frac{1}{|M_{\hat{b}_i\hat{b}_j}|} \sum_{(x,y)}^{M_{\hat{b}_i\hat{b}_j}} cos^{-1}\left(\frac{\hat{b}_i^{xy} \cdot w_i(\hat{b}_j)^{xy}}{\|\hat{b}_i^{xy}\| \|w_i(\hat{b}_j)^{xy}\|} \right),
\end{equation}
where $(x,y) \in M_{\hat{b}_1\hat{b}_2}$ are the pixel locations in $\hat{b}_i$ that have a matching pixel in $\hat{b}_j$. $\hat{b}_i^{xy}$ is the RGB vector at this location. $w_i(\cdot)$ is the warping operator into $\hat{b}_i$'s pixel space. Note that the angle between vectors $u,v$ can be computed by $cos^{-1}((u\cdot v) / (\|u\| \|v\|))$.
This enforces consistent hue in corresponding locations while allowing varying brightness.

\begin{equation} \label{eq:total_loss}
    L_{total} = L_{translation} + \lambda L_{vc}
\end{equation}
Equation~\ref{eq:total_loss} shows the final loss function. To avoid an imbalance between domains $A$ and $B$, $L_{translation}$ is not enforced on $\hat{b}_j$ and errors from $L_{vc}$ are only backpropagated through $\hat{b}_i$ and not $\hat{b}_j$. $\lambda$ is initialized with $0$ and set to $20$ after $10^4$ training steps to avoid forcing consistency on unrefined translations in early stages of training. Complete training details can be found in the supplementary.

\subsection{Data}
For the domain of real images $B$, we collected 28 recordings of robotic, abdominal surgeries from the University Hospital Carl Gustav Carus Dresden and manually selected sequences which contain views of the liver. The institutional review board approved the usage of this data. Frames were extracted at 5fps, resulting in a total of 13,334 training images. During training, images are randomly resized and cropped to size 256x512.

For the simulated domain $A$, we built seven artificial abdominal 3D scenes in Blender containing liver, liver ligament, gallbladder, abdominal wall and fat/stomach. The liver meshes were taken from a public dataset (3D-IRCADb 01 data set, IRCAD, France) while all other structures were designed manually. For each scene, we generated 3,000 random views of size 256x512, resulting in a total of 21,000 training views. To evaluate temporal consistency, we manually created seven 20-second sequences at 5fps which pan over each scene with varying viewpoints and distances.

\begin{figure*}[t]
\begin{center}
\includegraphics[width=\linewidth, trim= .5cm 1.4cm 2cm .8cm, clip]{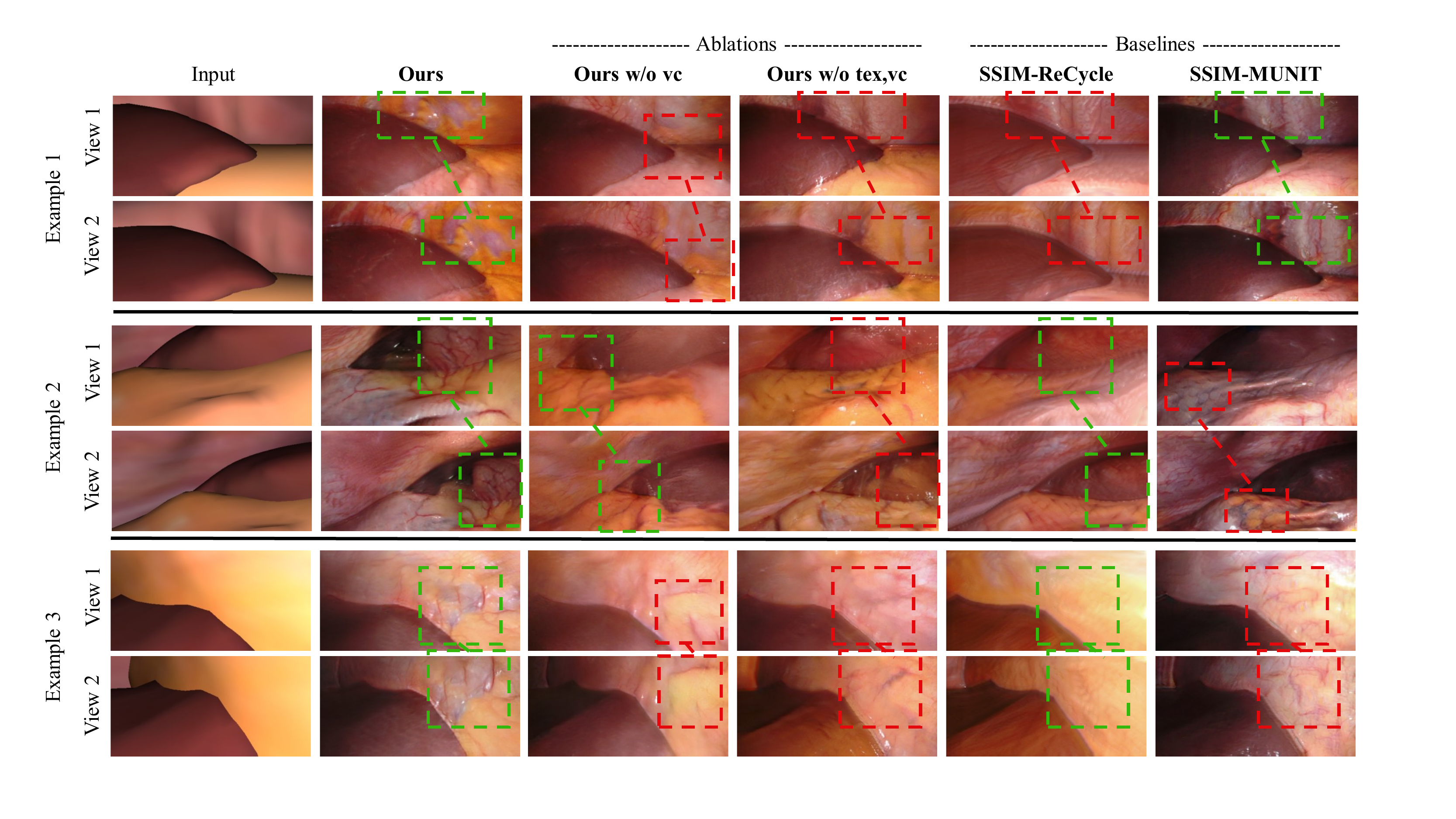}
\end{center}
\caption{Qualitative comparison: View-consistent areas are marked {\color{green}green}, inconsistent ones {\color{red}red}. Our model can render fine-grained details consistently across views. \emph{SSIM-ReCycle} often produces consistent outputs but lacks detail and realism. \emph{SSIM-MUNIT} produces realistic but flickering results. Quality and consistency can best be judged in videos at \url{http://opencas.dkfz.de/video-sim2real}.}
\label{fig:comparison}
\end{figure*}

\begin{table*}[t]
\begin{center}
\begin{tabular}{|l|c|c|c|c|c|c|c|c|}
\hline
Method & Data & \multicolumn{3}{|c|}{Realism} & \multicolumn{4}{|c|}{Temp. Consistency}\\
& & FID $\downarrow$ & KID $\downarrow$ & Dice $\uparrow$ & OF $\downarrow$ & ORB-1 $\uparrow$ & ORB-5 $\uparrow$ & ORB-10 $\uparrow$ \\
& & & & \% & & \% (\# per pair) & \% (\# per pair) & \% (\# per pair) \\
\hline\hline
SSIM-MUNIT~\cite{pfeiffer2019generating} & img & 28.3 & .0132 & {\bf59.2} & 8.64 & 60.5\% (32.5) & 36.1\% (15.7) & 19.1\% (7.0) \\
ReCycle~\cite{bansal2018recycle} & vid & 61.5 & .0454 & 40.7 & 8.89 & 69.6\% (16.3) & 43.9\% (7.0) & 23.5\% (2.7) \\
SSIM-ReCycle & vid & 80.6 & .0622 & 50.9 & 8.75 & 88.9\% (13.3) & 67.4\% (6.2) & 43.2\% (2.8) \\
OF-UNIT & vid & {\bf26.8} & .0125 & 57.7 & 8.53 & {\bf93.5\%} (32.4) & 59.0\% (11.1) & 30.7\% (4.4) \\
OF-UNIT (revisit) & vid & - & - & - & 8.91 & 69.9\% (15.7) & 43.8\% (7.3) & 24.7\% (3.4) \\
\hline
Ours w/o tex,vc & img & 27.3 & {\bf.0114} & 56.8 & {\em8.43} & 81.7\% (31.2) & 51.3\% (13.3) & 29.5\% (6.2) \\
Ours w/o vc & vid & 27.0 & .0134 & 55.2 & 8.35 & 88.3\% (27.9) & 66.8\% (14.6) & 44.5\% (7.5) \\
Ours & vid & {\bf26.8} & .0124 & 57.1 & {\bf7.62} & 91.8\% ({\bf49.7}) & {\bf73.0\%} (\bf{27.2}) & {\bf49.6\%} (\bf{13.9}) \\
\hline
\end{tabular}
\end{center}
\caption{Quantitative results with best scores printed {\bf bold}. For metrics \emph{ORB-1}, \emph{ORB-5} and \emph{ORB-10}, we report the accuracy of feature matches and the total number of correct matches per image pair, indicating both consistency as well as level of detail.}
\label{tbl:results}
\end{table*}

\section{Experiments}
To establish that our method produces both realistic and long-term consistent outputs, we need to evaluate the quality of individual images as well as consistency between consecutive or non-consecutive frames. Thus, we establish several baselines and evaluate them using various metrics. We place a special focus on both detailed and temporally consistent translation, since correct re-rendering of details such as vessels is crucial for obtaining realistic videos.

\subsection{Baselines}

{\bf SSIM-MUNIT:} This is Pfeiffer \etal's~\cite{pfeiffer2019generating} model for surgical image translation trained on our dataset of real and synthetic images $b$ and $a^{ref}$.
It corresponds to our image translation module but with added styles and noise injected into generator input. We remove these components in our model since they are disadvantageous for view consistency.

{\bf ReCycle and SSIM-ReCycle:} We compare to Bansal \etal's unpaired video translation approach ReCycle-GAN~\cite{bansal2018recycle} which is trained on triplets of consecutive video frames to maintain temporal consistency. We use the variant with additional non-temporal cycles
(\url{https://github.com/aayushbansal/Recycle-GAN}).
Additionally, we implement a variant with MS-SSIM loss for label preservation.

{\bf OF-UNIT:} State-of-the-art unpaired video translation models condition the generator on translations from previous time-steps to ensure short-term temporal consistency. Many methods thereby warp the previous image by estimating optical flow (OF) and achieve incremental improvement through better OF estimation~\cite{chen2019mocycle,chu2020learning,park2019preserving}. We argue, however, that even perfect OF is not enough for long-term consistency and can even have detrimental effects as we show later. To demonstrate this, we build a variant of our model which uses ground-truth OF to warp the previous translation, \ie it can potentially produce perfect transitions between frames. We replace the input $(a^{tex},a^{ref})$ of the encoder $E_A$ with $(w(\hat{b}_{prev}),a^{ref})$, where $\hat{b}_{prev}$ is the generated frame of the previous time step and $w$ is the perfect warping operator using ground-truth optical flow. Thus, \emph{OF-UNIT} serves as an upper-bound for the state of the art in unpaired video translation, where OF has to be estimated and is therefore imperfect.

{\bf Ours w/o vc and Ours w/o tex, vc:} Finally, we ablate our model by removing first the view-consistency loss and then also neural textures. The second model corresponds to SSIM-MUNIT without styles or noise in the generator.

\subsection{Metrics}
{\bf Realism:} We compare the realism of models through the commonly used metrics \emph{Frechet Inception Distance (FID)}~\cite{heusel2017gans} and \emph{Kernel Inception Distance (KID)}~\cite{binkowski2018demystifying} for which we sample 10,000 random images from the set of real and generated training images, each. Further, we train a U-Net variant for liver segmentation on a dataset of 405 laparoscopic images from 5 patients and report the \emph{Dice} score when evaluated on all 21,000 generated images. This metric measures both realism and label preservation.

{\bf Temporal Consistency:} We introduce two metrics to evaluate the temporal consistency of the sequences generated from each scene. Firstly, we measure the mean absolute error for the estimated optical flow \emph{OF} of consecutive translated frames $\hat{b}_t,\hat{b}_{t+1}$ and their corresponding simulated reference images $a_t^{ref}, a_{t+1}^{ref}$ by $mean(|OF(a_t^{ref}, a_{t+1}^{ref}) - OF_{GF}(\hat{b}_t,\hat{b}_{t+1})|)$ where $OF(a_t^{ref}, a_{t+1}^{ref})$ is the ground truth optical flow of the synthetic scene and $OF_{GF}(\hat{b}_t,\hat{b}_{t+1})$ is the optical flow estimated by the Gunnar-Farneback method~\cite{farneback2003two} on the generated frames. As argued by Chu \etal~\cite{chu2020learning}, this is better than the more common RGB error on warped images, since the latter favors blurry sequences. Secondly, the metrics \emph{ORB-1, ORB-5 and ORB-10} measure how consistently image features are rendered. For \emph{ORB-1}, we compute all ORB feature~\cite{rublee2011orb} matches in consecutive frames and determine whether the matched feature points correspond the the same 3D location. We report the accuracy of matches as well as the average number of correct matches per image pair. A blurry but consistent sequence might yield a high accuracy, so the number of matches gives additional information on how detailed the results are. A match is considered correct if its distance is smaller than 1mm in the underlying 3D scene. To investigate consistency beyond consecutive frames, we do the same with pairs that are 5 and 10 frames apart (\ie 1 and 2 sec.) as \emph{ORB-5} and \emph{ORB-10}.

{\bf Time-independence:} Finally, we show the pitfalls of previous approaches which condition on previous time steps. We extend each test sequence by running it first forward and then backward such that each view is visited twice with varying temporal distance. I.e. given a sequence $1,\dots,T$, we extend it to $1,\dots T,T,\dots,1$ similar to Mallya \etal's~\cite{mallya2020world} evaluation. We then compute the same metrics \emph{OF} and \emph{ORB-1}. But instead of comparing frame $t$ to its successor $t+1$, we use the time point in the extended sequence which corresponds to its successor, namely $2T-t$. For all methods except \emph{OF-UNIT}, this is equivalent to the original metric since they depend only on the current view. Analogously for \emph{ORB-5} and \emph{ORB-10}, we compare to time points $2T-t-4$ and $2T-t-9$, respectively. We denote these experiments as \emph{OF-UNIT (revisit)}.

\section{Results}

\begin{figure}
\begin{center}
\includegraphics[width=.9\linewidth, trim=1cm 1.2cm 1cm 1.3cm, clip]{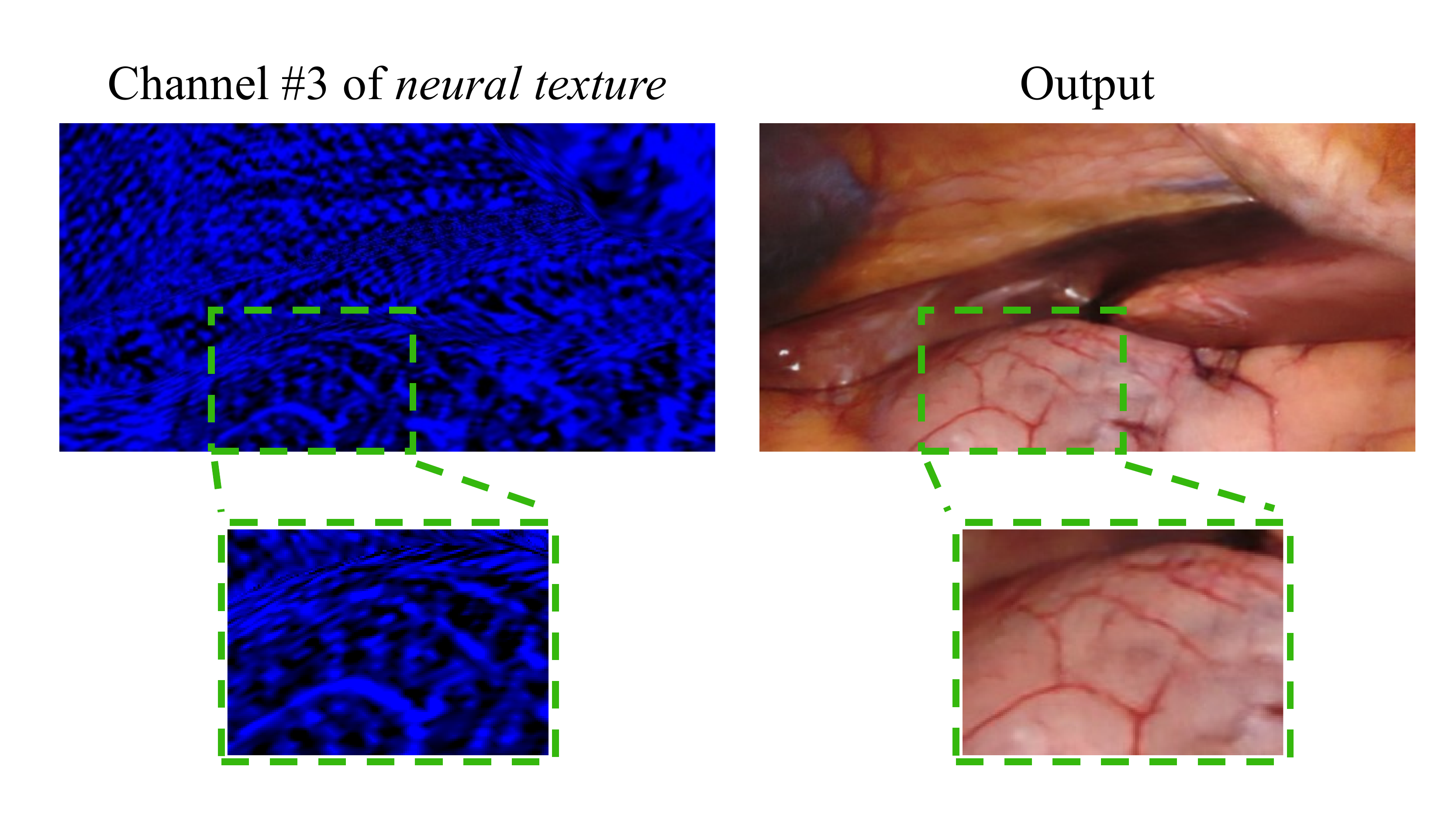}
\end{center}
   \caption{Details are stored in our \emph{neural textures}. We found the 3rd feature channel often to correspond to vessels.}
\label{fig:texture}
\end{figure}

\subsection{Realism}
Table \ref{tbl:results} shows our model achieves similar FID and KID scores as image-based approaches (\emph{SSIM-UNIT} and \emph{Ours w/o tex,vc}) while strongly outperforming video-based methods \emph{ReCycle} and \emph{SSIM-ReCycle}. We hypothesize that their temporal cycle-loss favors blurry images since they are easier to predict for the temporal prediction model. Fig. \ref{fig:comparison} supports this hypothesis as our and image-based models show more detailed and realistic translations than \emph{SSIM-ReCycle}. For \emph{OF-UNIT}, similar realism scores to ours are expected, since it uses the same translation module.

\begin{figure}
\begin{center}
\includegraphics[width=\linewidth, trim=6.5cm 4.5cm 9.5cm 4.7cm, clip]{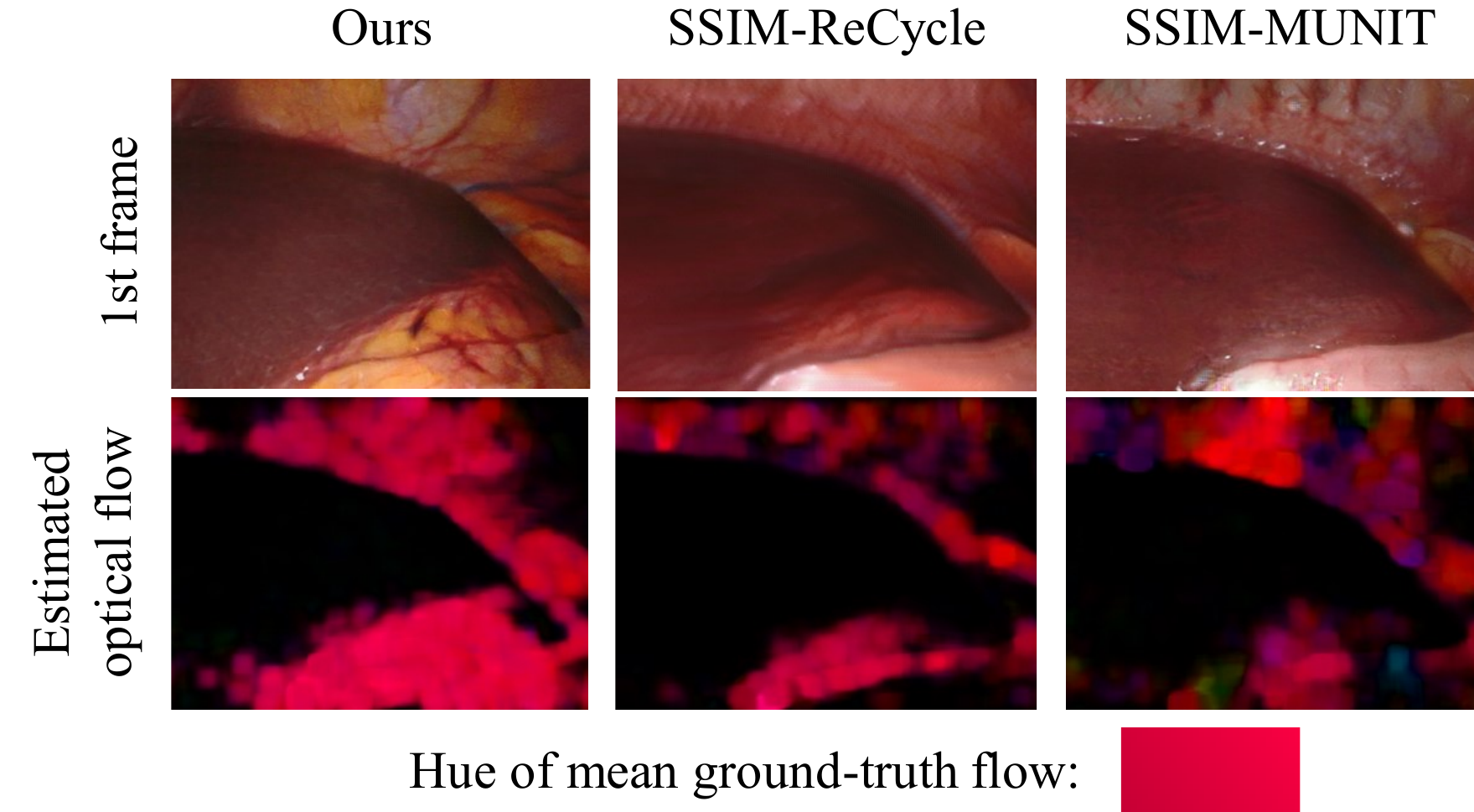}
\end{center}
   \caption{Estimated optical flow in a scene with camera motion (where hue indicates the direction of the flow). In our results, consistent motion is detected on textured surfaces while blurriness or flickering lead to poor flow estimates in other models.}
\label{fig:of}
\end{figure}

Further, we evaluate a pretrained liver segmentation network on the generated data. Again, our model yields comparable results to image-based methods while outperforming \emph{ReCycle} and \emph{SSIM-ReCycle}. This indicates that our results are not only realistic but content of the simulated domain is also translated correctly.
The gap between \emph{ReCycle} and \emph{SSIM-ReCycle} additionally shows the importance of the MS-SSIM loss for label-preservation.
Example 2 in Fig. \ref{fig:comparison} shows a failure case of our model where a stomach-like texture with vessels is rendered on the liver. Introducing \emph{neural textures} supposedly improves the sharpness and level of detail in translations but increases the model's freedom to change content in the scene. The quantitative results, however, suggest that this is only a minor effect.

\subsection{Temporal Consistency}

Using the established ORB feature detector, we evaluate how consistently visual features are re-rendered in following frames of generated video sequences. We report how often detected feature matches are correct as well as the number of correct matches per pair of frames. For neighboring frames, our model achieves an accuracy of $91.8\%$, outperforming all baselines except \emph{OF-UNIT}. However, this is not surprising since the latter uses the perfectly warped previous frame as input. For larger frame distances, however, our model outperforms \emph{OF-UNIT}, showing its superiority w.r.t. long-term consistency. Additionally, the absolute number of correct matches per image pair is drastically higher than in \emph{OF-UNIT} and other models even for neighboring frames. This indicates that our \emph{neural textures} not only enable consistent translation but also encourage more detailed rendering. Fig. \ref{fig:comparison} shows several translated views with detailed as well as consistent textures. In Fig. \ref{fig:texture}, we show how the location of vessels is stored in the \emph{neural texture}.

We observe that other methods fail to generate detailed as well as temporally consistent sequences. While \emph{SSIM-MUNIT} produces detailed translations (indicated by the high number of matches), it achieves the lowest accuracies. Oppositely, video-based \emph{ReCycle} and \emph{SSIM-ReCycle} produce more consistent but less detailed renderings, indicated by their high accuracy but low number of correct matches.

\begin{figure}
\begin{center}
\includegraphics[width=.9\linewidth, trim=8cm 5cm 8cm 4.15cm, clip]{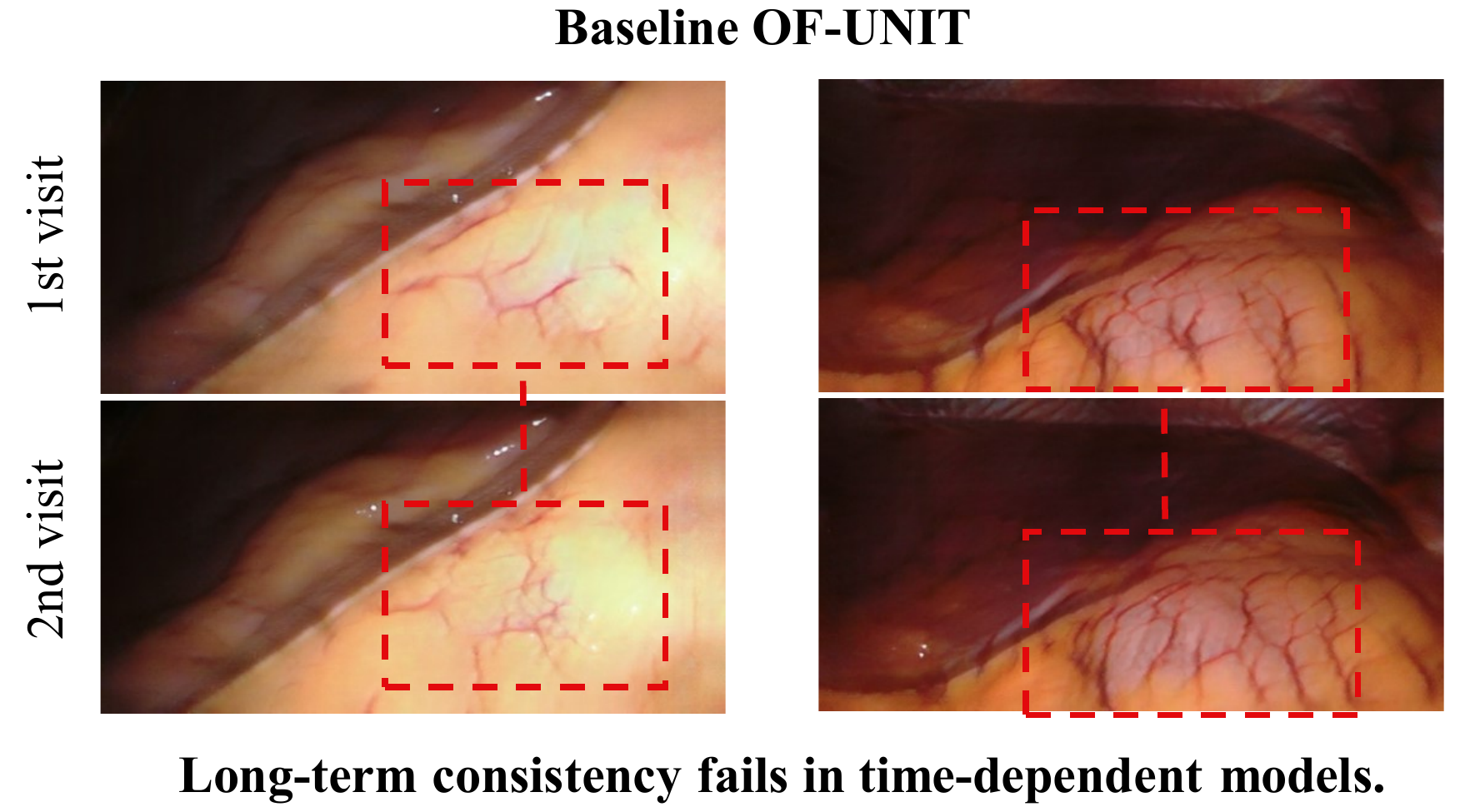}
\end{center}
   \caption{When revisiting a previous view, time-dependent models such as OF-UNIT fail to render textures consistently. Our model maintains consistency independently of the duration between visits by storing information in texture-space.}
\label{fig:revisited}
\end{figure}

Note that \emph{SSIM-MUNIT} induces flickering since noise is injected into the generator. Temporal consistency can already be strongly improved by removing this component (\emph{Ours w/o tex,vc}). Adding \emph{neural textures} without enforcing view-consistency (\emph{Ours w/o vc}) further improves results.

Evaluating temporal consistency through optical flow (OF) supports our previous findings. This metric measures both temporal consistency as well as level of detail, since Gunnar-Farneback flow often fails on smooth surfaces. Image- and other video-based methods yield high errors, since the former tend to produce detailed but flickering sequences, while the latter often generate blurry but consistent views (Fig. \ref{fig:of}). By learning textures in 3D space, our model achieves both detailed and consistent renderings.

\subsection{Time-independence}
We have seen that the time-dependent baseline \emph{OF-UNIT} achieves very consistent transitions between frames and still achieves respectable results for larger frame distances. However, if the second frame is replaced with the same view revisited at a later point of the sequence, then performance drastically degrades. This is because the model does not have the capacity to \emph{remember} the appearance of areas which have left the field of view. It even underperforms compared to its unconditioned variant \emph{Ours~w/o~tex,vc}. We hypothesize that dependence on the previous trajectory actually encourages appearance changes over time (Fig \ref{fig:revisited}). We believe time-independence is therefore an important feature for achieving long-term consistency, even in non-static scenes. With our approach, moving objects as well as deformations can potentially be handled by moving or deforming the neural texture accordingly.

\begin{figure}
\begin{center}
\includegraphics[width=.85\linewidth, trim=1.2cm 6.7cm 1.5cm 1.5cm, clip]{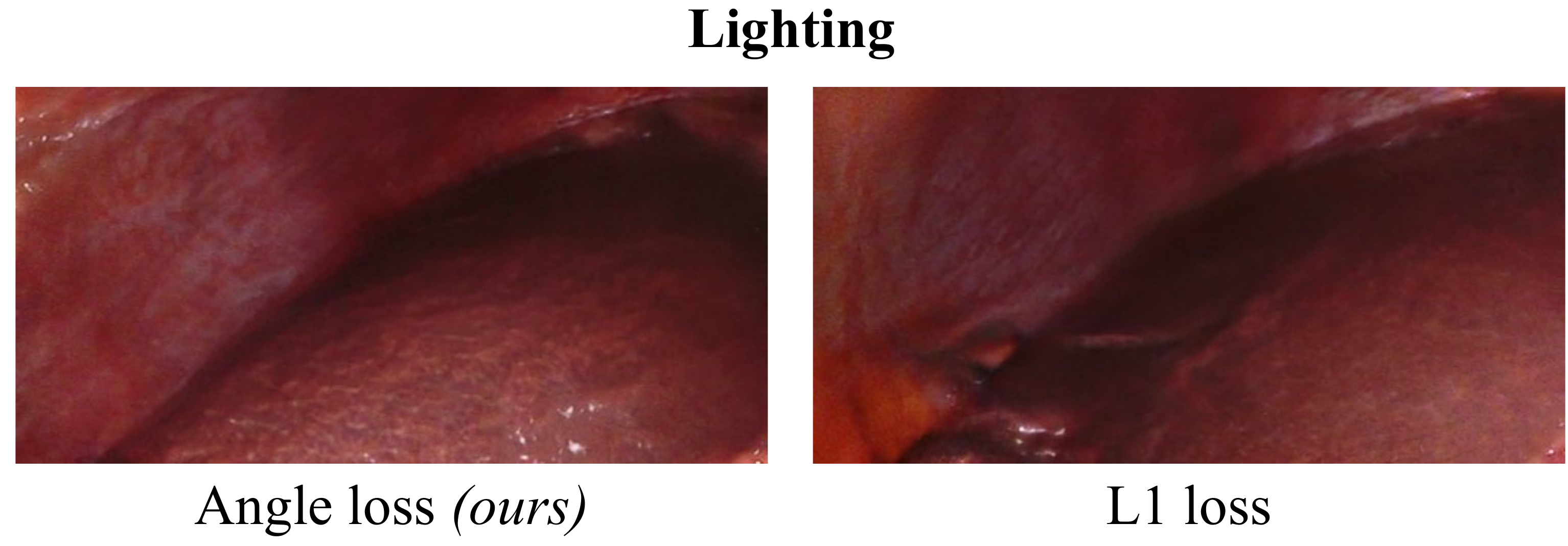}
\end{center}
   \caption{Our angle loss allows the translation module to adjust brightness of areas according to the current view. In real images, the center is often brightest since the light source is mounted on the camera.}
\label{fig:angle}
\end{figure}

\subsection{Lighting-invariant View Consistency}
We proposed an angle-based loss for view consistency which only keeps the hue of corresponding areas consistent. Fig. \ref{fig:angle} shows that our angle loss allows for more realistic lighting since the translation module can change brightness according to the current view. On the other hand, an L1 loss enforces static brightness from arbitrary viewpoints. This results in incorrect lighting like in the left image where the light appears to come from the bottom right. More examples can be found in the supplementary material.

\section{Conclusion}
We combine neural rendering with unpaired image translation from simulated to photorealistic videos.
We target surgical applications where labeled data is often limited and realistic but simulated evaluation environments are especially relevant.
Through extensive evaluation and comparison to related approaches, we show that our results maintain the realism of image-based approaches while outperforming video-based methods w.r.t. temporal consistency. We show that optical flow is consistent with the underlying simulated scene and that our model can render fine-grained details such as vessels consistently from different views.
Also, data generation can easily be scaled up by adding more simulated scenes.
A crucial observation about the model is that it leverages the rich information contained in the simulated domain while requiring only an unlabeled set of images on the real domain. This way, consistent and label-preserving data can be generated without limiting its relevance for real-world applications.
Specifically, ground truth which would be unobtainable in surgical settings can be generated (\eg depth, optical flow, point correspondences).
This work is a step towards more expressive simulated environments for \eg surgical assistance systems, robotic applications or training aspiring surgeons.
While we focus on surgical applications (where access to labeled data is especially restricted), the model can potentially be used for any setting with a simulated base for translation.

\paragraph*{Acknowledgements} Funded by the German Research Foundation (DFG, Deutsche Forschungsgemeinschaft) as part of Germany’s Excellence Strategy – EXC 2050/1 – Project ID 390696704 – Cluster of Excellence “Centre for Tactile Internet with Human-in-the-Loop” (CeTI) of Technische Universität Dresden.

{\small
\bibliographystyle{ieee_fullname}
\bibliography{iccv2021_7515}
}

\clearpage

\appendix

\section{Implementation details}

\subsection{Neural Texture \& Projection Mechanism}
Here, we go into more detail on the projection mechanism between texture and pixel space and from which we obtain the image-sized feature map $a^{tex}_i$:
\begin{equation} \label{eq:project}
    \bm{a^{tex}_i} = project(tex,view_i)
\end{equation}
In the following, we describe how the projection works for a single pixel $(x,y)$ of a given $view_i$.

First, a ray is cast through the pixel to the surface of the $3D$ scene. We obtain a surface coordinate $s \in \mathbb{R}^3$ and its corresponding object $o \in \{\text{'liver'},\text{'gallbladder'},\text{'ligament'},\text{'abdominal wall'},\text{'fat'}\}$.
\begin{equation}
    ray(view_i,x,y) = s,o
\end{equation}
From the surface point $s$, we find the corresponding texel coordinates for each of the three texture planes which face the surface point (\ie where the dot-product of normals is positive). Assuming surface and texture normals are never exactly orthogonal in practice, $tri(s)$ always returns three sets of texture plane coordinates.
\begin{equation}
\begin{aligned}
    tri(s) &= \bigl\{proj_p(s)|p\in \{1..6\} \land (n_s^T \cdot n_p)>0\bigl\}\\
    &= \{(x_{p_1},y_{p_1}),(x_{p_2},y_{p_2}),(x_{p_3},y_{p_3})\}
\end{aligned}
\end{equation}
where $x_{p_i},y_{p_i}$ are the texel coordinates of $s$ projected onto texture plane $p_i$ and transformed into texel space.

To obtain the final, projected features in pixel space, we compute a weighted average of the texture features $tex[o,p,x_p,y_p]$ from the three planes.
\begin{equation}
    \bm{a_i^{tex}[x,y]} = \sum_{x_p,y_p}^{tri(s)} w_{tri}(s,p) \cdot tex[o,p,x_p,y_p]
\end{equation}
\begin{equation}
    w_{tri}(s,p) = (n_s^T \cdot n_p)^2
\end{equation}
Features are weighted according to the squared dot-product of surface and texture normal, \ie the texture plane which faces the surface point most, contributes most to the final feature vector.
Note that $\|n_s\| = 1$ and $n_p$ are orthogonal one-hot or negative one-hot vectors. Hence, the sum of weights is always $n_{s,x}^2+n_{s,y}^2+n_{s,z}^2 = \|n_s\|^2 = 1$.

To obtain texture features $tex[o,p,x_p,y_p]$ from continuous coordinates $x_p,y_p$, we use bilinear interpolation:
\begin{equation}
\begin{aligned}
    tex[o,p,x_p,y_p] \span \span \span \span \span & & \\
    &=& (1- &\Delta x)&(1-&\Delta y)& &tex\bigl[o,p,\lfloor x_p\rfloor,\lfloor y_p\rfloor\bigr]\\
    &+& (1- &\Delta x)& &\Delta y& &tex\bigl[o,p,\lfloor x_p\rfloor,\lceil y_p\rceil\bigr]\\
    &+& &\Delta x&(1- &\Delta y)& &tex\bigl[o,p,\lceil x_p\rceil,\lfloor y_p\rfloor\bigr]\\
    &+& &\Delta x& &\Delta y& &tex\bigl[o,p,\lceil x_p\rceil,\lceil y_p\rceil\bigr]
\end{aligned}
\end{equation}
where $\Delta x = x_p - \lfloor x_p \rfloor$ and $\Delta y = y_p - \lfloor y_p \rfloor$.

\subsection{Warping}
To enforce the view-consistency loss $L_{vc}$, we have to warp the translated image $\hat{b}_j$ of $view_j$ into the pixel space of $view_i$. To this end, we define the warping operator $w_i(\cdot)$ to obtain the warped image
\begin{equation}
    \bm{w_i(\hat{b}_j)}.
\end{equation}
To map pixel coordinates $x,y$ from $view_j$ to $view_i$, we first use the ray function $ray(view_j,x,y)$ from $view_j$ to obtain a $3D$ surface point $s=(s_x,s_y,s_z)$. Using the $view_i$'s intrinsic and extrinsic camera parameters, $s$ is projected back into $view_i$ and we obtain the corresponding pixel coordinates $x_i,y_i$.
\begin{equation}
    \begin{pmatrix}
    x_i\\
    y_i\\
    1
    \end{pmatrix}
    = C_i \cdot RT_i \cdot
    \begin{pmatrix}
    s_x\\
    s_y\\
    s_z\\
    1
    \end{pmatrix},
\end{equation}
where $C_i \in \mathbb{R}^{3\times 3}$ is $view_i$'s intrinsic matrix with focal lengths $f_u,f_v$ and principal point $(u_0,v_0)$ and $RT_i \in \mathbb{R}^{3\times 4}$ is its extrinsic matrix with rotational and translational parameters $\{r_{..}\},\{t_.\}$. All parameters can be extracted from the simulated scene.
\begin{equation}
    C_i = 
    \begin{pmatrix}
    f_u & 0 & u_0\\
    0 & f_v & v_0\\
    0 & 0 & 1
    \end{pmatrix}
\end{equation}
\begin{equation}
    RT_i =
    \begin{pmatrix}
    r_{00} & r_{01} & r_{02} & t_0\\
    r_{10} & r_{11} & r_{12} & t_1\\ 
    r_{20} & r_{21} & r_{22} & t_2\\ 
    \end{pmatrix}
\end{equation}
Using this correspondence between pixel coordinates of both views, $RGB$ values can now be mapped into the warped image:
\begin{equation}
    \bm{w_i(\hat{b}_j)[x_i,y_i]} = 
    \begin{cases}
    \hat{b}_j[x,y] &\text{if } \neg occl_i(s,s_i)\\
    0, &\text{else.}
    \end{cases}
\end{equation}
where $s,\_ = ray(view_j,x,y)$ is the $3D$ surface point we want to warp and $s_i,\_ = ray(view_i,x_i,y_i)$ is the surface point we obtain if a ray is cast into the surface from the warped location in the target view $i$.
Note that that we do not map surface points from $view_j$ if they are occluded in $view_i$. To this end, we define the boolean occlusion function $occl_i(s_1,s_2)$:
\begin{equation}
    occl_i(s_1,s_2) = depth_i(s_1) > (depth_i(s_2) + \epsilon),
\end{equation}
where $depth_i(s)$ is the depth of a surface point $s$ from $view_i$. To account for inaccuracies, we say it is an occlusion if the former depth is larger than the latter by a margin of $\epsilon=1$mm in the blender scene.

\subsection{Network architectures}
We repurpose MUNIT's~\cite{huang2018multimodal} network architectures for encoders $E_A,E_B$, decoders $G_A,G_B$ and discriminators $D_A,D_B$, but remove styles from the encoders and decoders. \emph{AdaIN} layers are replaced by instance normalization layers.

\subsection{Translation}
In each training iteration, two simulated views and one real image are sampled:
\begin{equation}
    \begin{pmatrix}
    a^{tex}_i\\
    a^{ref}_i
    \end{pmatrix},
    \begin{pmatrix}
    a^{tex}_j\\
    a^{ref}_j
    \end{pmatrix} \in A,
    \qquad
    b \in B
\end{equation}
Equations \ref{eq:translation_first} to \ref{eq:translation_last} show all translations and reconstructions performed during each training step. Note that the second view $j$ from domain $A$ is only translated to domain $B$ and not used for reconstructions or cycles.

\paragraph{Domain A:}
\begin{equation} \label{eq:translation_first}
    \begin{pmatrix}
    a^{tex}_i\\
    a^{ref}_i
    \end{pmatrix}
    \xrightarrow{E_A}
    c^{a_i}
    \xrightarrow{G_B}
    \hat{b}_i
    \xrightarrow{E_B}
    c^{a_i}_{rec}
    \xrightarrow{G_A} 
    \begin{pmatrix}
    a_{i,cyc}^{tex}\\
    a_{i,cyc}^{ref}
    \end{pmatrix}
\end{equation}
\begin{equation}
    \begin{pmatrix}
    a^{tex}_i\\
    a^{ref}_i
    \end{pmatrix}
    \xrightarrow{G_A \circ E_A}
    \begin{pmatrix}
    a_{i,rec}^{tex}\\
    a_{i,rec}^{ref}
    \end{pmatrix}
\end{equation}

\begin{equation}
    \begin{pmatrix}
    a^{tex}_j\\
    a^{ref}_j
    \end{pmatrix}
    \xrightarrow{G_B \circ E_A}
    \hat{b}_j
\end{equation}

\paragraph{Domain B:}
\begin{equation}
    b
    \xrightarrow{E_B}
    c^b
    \xrightarrow{G_A}
    \begin{pmatrix}
    \hat{a}^{tex}\\
    \hat{a}^{ref}
    \end{pmatrix}
    \xrightarrow{E_A}
    c^b_{rec}
    \xrightarrow{G_B} 
    b_{cyc}
\end{equation}

\begin{equation} \label{eq:translation_last}
    b
    \xrightarrow{G_B \circ E_B}
    b_{rec}
\end{equation}

\begin{figure*}
\begin{center}
\includegraphics[width=.95\linewidth, trim= 2cm 0cm 3cm 0cm, clip]{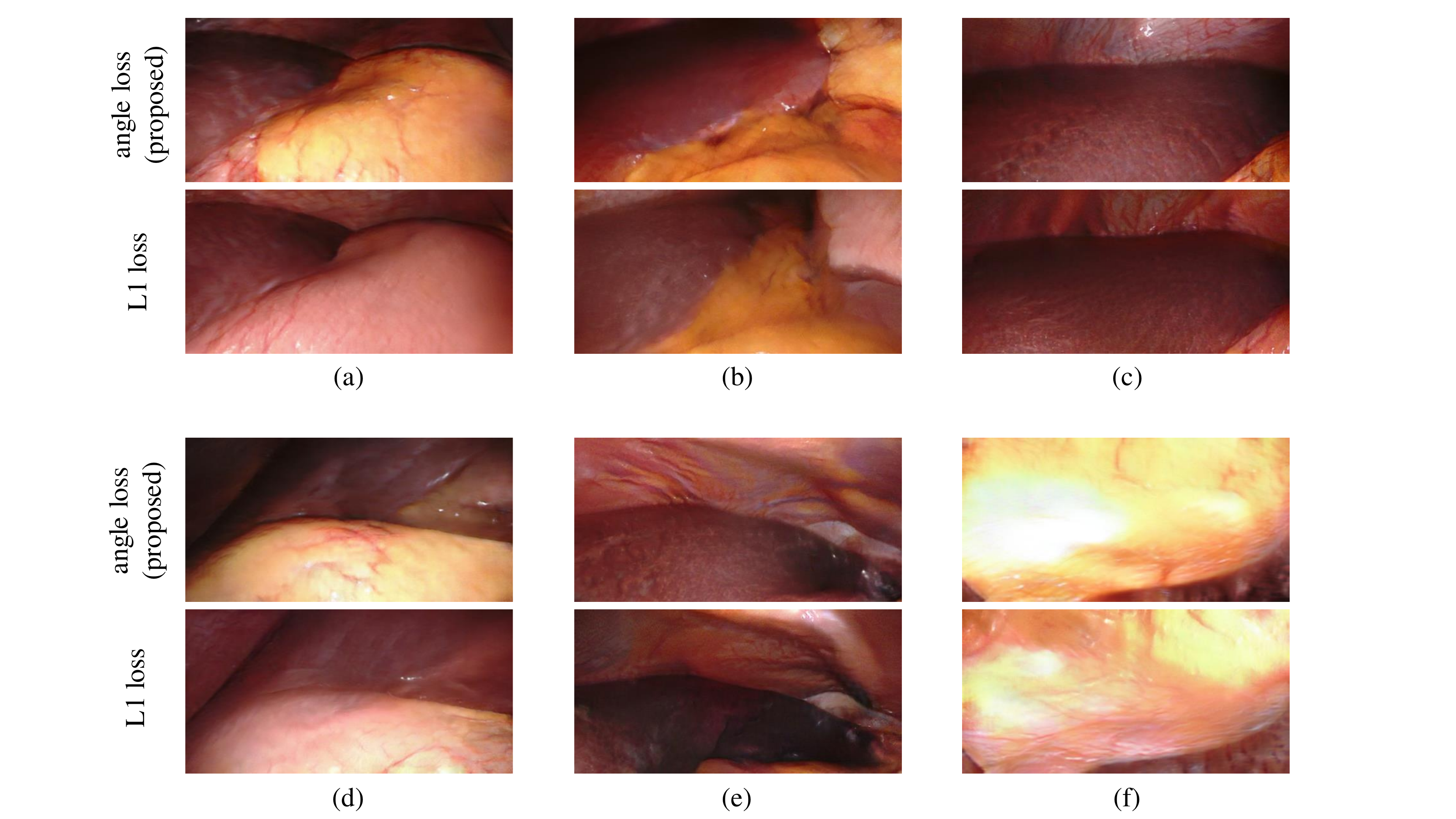}
\end{center}
\caption{More examples of our proposed angle-based loss vs. an L1 loss. While the angle-based loss allows close and central parts of the image to be brightened, the L1 loss leads to either uniform brightness (\emph{(a)-(d)}) or unrealistic lighting conditions (\emph{(e)} and \emph{(f)}).}
\label{fig:angle_suppl}
\end{figure*}

\subsection{Losses}

\paragraph{Generator/texture update:} Encoders $E_A,E_B$, decoders $G_A,G_B$ and neural textures $tex$ are learned by minimizing Equation~\ref{eq:loss_total}.
\begin{equation} \label{eq:loss_total}
    L_{total} = L_{translation} + \lambda L_{vc}
\end{equation}

\begin{equation}
\begin{split}
    L_{translation} = L_{adv} + L_{cyc} + L_{rec} + L_{c} + L_{ssim},
\end{split}
\end{equation}

\begin{equation}
    L_{adv} =
    \norm{
        D_A
        \begin{pmatrix}
        \hat{a}^{tex}_i\\
        \hat{a}^{ref}_i
        \end{pmatrix}
        -1
    }^2_2
    +
    \norm{D_B(\hat{b}) -1}^2_2
\end{equation}

\begin{equation}
    L_{cyc} =
    \lambda_{cyc}
    \biggl(
    \norm{
        \begin{pmatrix}
        a^{tex}_i - a^{tex}_{i,cyc}\\
        a^{ref}_i - a^{ref}_{i,cyc}
        \end{pmatrix}
    }_1
    +
    \norm{b - b_{cyc}}_1
    \biggr)
\end{equation}

\begin{equation}
    L_{rec} =
    \lambda_{rec}
    \biggl(
    \norm{
        \begin{pmatrix}
        a^{tex}_i - a^{tex}_{i,rec}\\
        a^{ref}_i - a^{ref}_{i,rec}
        \end{pmatrix}
    }_1
    +
    \norm{b - b_{rec}}_1
    \biggr)
\end{equation}

\begin{equation}
    L_{c} = \lambda_c (\norm{c^{a_i}-c^{a_i}_{rec}}_1 + \norm{c^b-c^b_{rec}}_1)
\end{equation}

\begin{equation}
\begin{aligned}
    L_{ssim} = & \lambda^{ab}_{ssim} \text{MS-SSIM}(gray(a^{ref}),gray(\hat{b}))\\
    + & \lambda^{ba}_{ssim} \text{MS-SSIM}(gray(b),gray(\hat{a}^{ref}))
\end{aligned}
\end{equation}

\begin{equation}
L_{vc} = \frac{1}{|M_{\hat{b}_i\hat{b}_j}|} \sum_{(x,y)}^{M_{\hat{b}_i\hat{b}_j}} cos^{-1}\left(\frac{\hat{b}_i^{xy} \cdot w_i(\hat{b}_j)^{xy}}{\|\hat{b}_i^{xy}\| \|w_i(\hat{b}_j)^{xy}\|} \right),
\end{equation}
We use $\lambda_{cyc}=10$, $\lambda_{rec}=10$, $\lambda_{c}=1$, $\lambda^{ab}_{ssim}=5$, $\lambda^{ba}_{ssim}=3$. For the Multi-Scale Structural Similarity (MS-SSIM) loss, we use Jorge Pessoa's implementation\footnote{\url{https://github.com/jorge-pessoa/pytorch-msssim}} with 5 scales and a window size of $11\times 11$. Similarity is only enforced on the brightness (gray values) of images. The weight for the view-consistency loss is initialized with $\lambda = 0$ and set to $20$ after $10$k iteration.

\paragraph{Discriminator update:} Discriminator networks $D_A,D_B$ are learned by minimizing:
\begin{equation}
\begin{aligned}
    L_{dis} =
    &
    \norm{
        D_A
        \begin{pmatrix}
        a^{tex}\\
        a^{ref}
        \end{pmatrix}
        -1
    }^2_2
    +
    \norm{D_B(b) -1}^2_2
    \\
    &
    \norm{
        D_A
        \begin{pmatrix}
        \hat{a}^{tex}\\
        \hat{a}^{ref}
        \end{pmatrix}
        -0
    }^2_2
    +
    \norm{D_B(\hat{b}) -0}^2_2
\end{aligned}
\end{equation}

\subsection{Training}
We train our model for 500,000 iterations and use the Adam optimizer with initial learning rates of $10^{-4}$ for network parameters and $10^{-3}$ for neural textures. Learning rates are halved every 100,000 iterations. For all experiments including ablations and baselines, we use a batch size of 1. Note, however, that the definition of a batch varies across methods. In our method we \eg sample one real and two simulated views per batch. In \emph{SSIM-MUNIT} only one image from each domain is sampled per batch, while for ReCycle or SSIM-Recycle 3 images for each domain are sampled.

\begin{figure*}
\begin{center}
\includegraphics[width=\linewidth, trim= 1cm 5cm 3.5cm 5cm, clip]{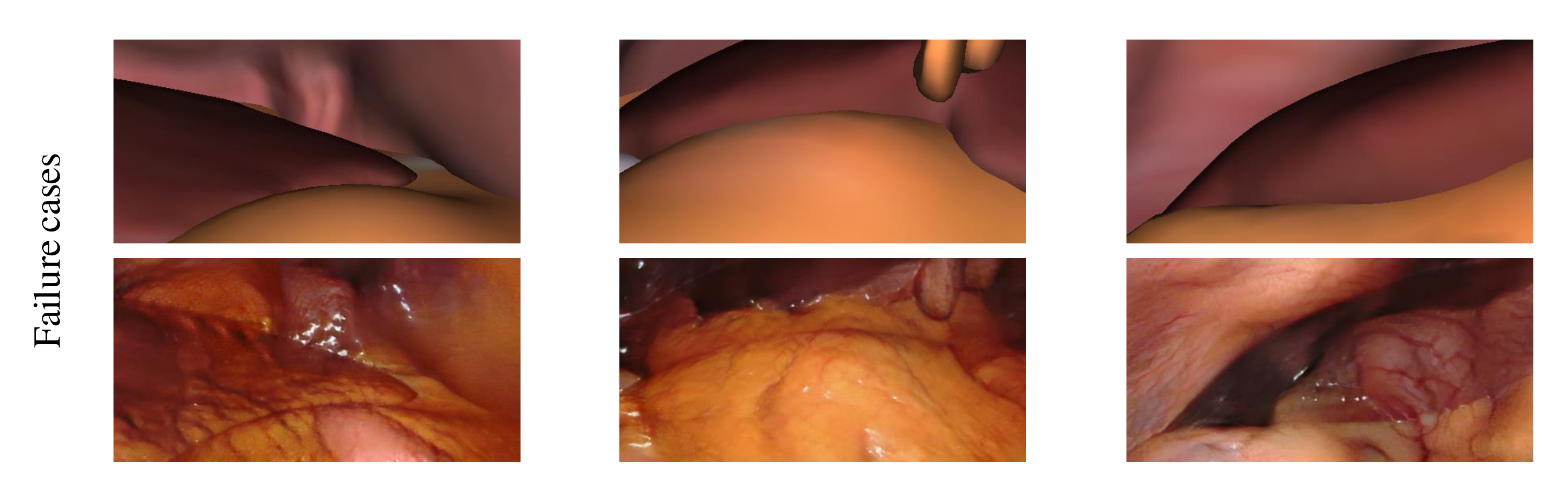}
\end{center}
\caption{Common failure cases include the rendering of fat or stomach texture on areas which are supposed to be liver (in the simulated scene).}
\label{fig:failue}
\end{figure*}

\subsection{Baselines}
\paragraph{SSIM-MUNIT:}
This model is trained for 370k iterations as in the original paper. For realism and label preservation experiments, we translate each training image once with randomly drawn styles and once with a style extracted from randomly drawn images. Test video sequences are translated with a randomly drawn but fixed style to guarantee a fair comparison with respect to temporal consistency.

\paragraph{ReCycle \& SSIM-ReCycle:}
Both models are trained for the suggested 40 epochs. Longer training times lead to degrading performance.
Triplets of simulated images are obtained through random, linear trajectories comparable to the motion in real sequences.

\paragraph{OF-UNIT:}
We train for $1$ Mio. iterations due to the stronger correlation between subsequent samples. We replace the view-consistency loss with an $L1$ loss and increase the weight $\lambda$ to $50$ since it gave better results. Intuitively, the angle loss is not required here since the views on which the loss is applied are consecutive frames and exhibit only minor lighting changes. Synthetic sequences are obtained from the triplets used for ReCycle.

\section{Additional Results}
Figure \ref{fig:angle_suppl} shows more translated views comparing our proposed angle-based view-consistency loss to a na\"ive $L1$ loss. Our proposed loss allows for view-dependent effects while an $L1$ loss actively discourages them.

\section{Limitations}

\paragraph{Realism of 3D Shapes}
The largest limitation of our current method is the lacking realism of the simulated 3D meshes. While liver meshes are obtained from real CT scans, all other organs were designed manually. Especially liver ligaments are difficult to model realistically. However, even the liver meshes often differ from realistic settings since intra-operative deformations (\eg by inflating the abdominal cavity) are different from ones observed during a CT scan. We manually deformed the liver meshes to resemble intra-operative shapes.

\paragraph{Freedom of Neural Textures}
Figure \ref{fig:failue} shows failure cases of our method. Sometimes the model misinterprets objects in the simulated scene and \eg renders fat or stomach textures on liver surfaces. Compared to previous work, this seems to happen slightly more frequently as indicated by our liver-segmentation experiments. We believe that neural textures give the model more freedom to interpret the simulated scene. Hence, enabling a higher level of detail and consistency comes with the trade-off of more misinterpretations.

\paragraph{Variability}
We propose a deterministic, style-less model since we believe that current state of the art for style-dependent translation (AdaIN~\cite{huang2017arbitrary}) is unsuitable and not theoretically sound in the video setting. To demonstrate this, we implement a variant of our model with AdaIN styles as used in MUNIT~\cite{huang2018multimodal} or SSIM-MUNIT~\cite{pfeiffer2019generating}.

\begin{figure*}
\begin{center}
\includegraphics[width=\linewidth, trim= 0cm 6cm 0cm 2cm, clip]{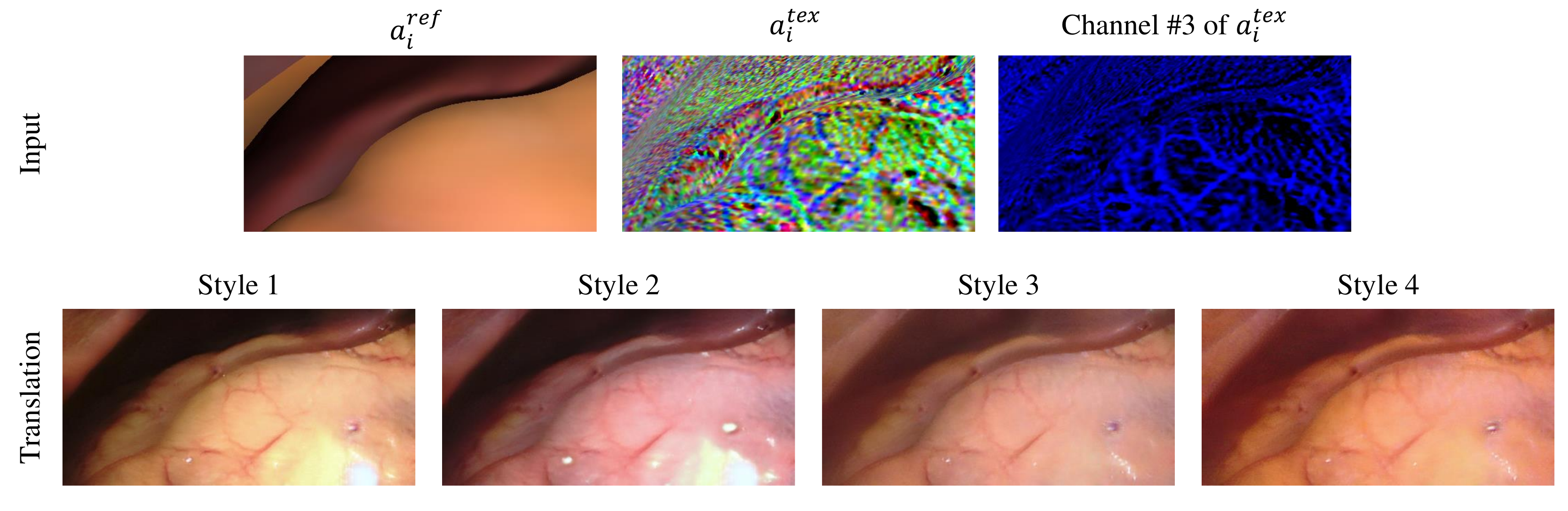}
\end{center}
\caption{Styles operate only at image-level but details such as vessels are stored in the neural textures (see \emph{Channel \#3 of} $a^{tex}_i$). Hence, the resulting variability is mostly restricted to color changes. It can be seen that vessels have identical locations in all samples. We believe research towards styles at texture-level would be a useful direction for future work.}
\label{fig:style_tex}
\end{figure*}

Firstly, styles operate at image level and do not introduce variability at texture level; \eg locations of vessels do not vary, but only their appearance (Fig.~\ref{fig:style_tex}). Hence, the variability is mostly limited to color changes but changing textures would be highly useful for creating diverse training and evaluation environments.

\begin{figure*}
\begin{center}
\includegraphics[width=\linewidth, trim= 2cm 1.5cm 2.5cm 2cm, clip]{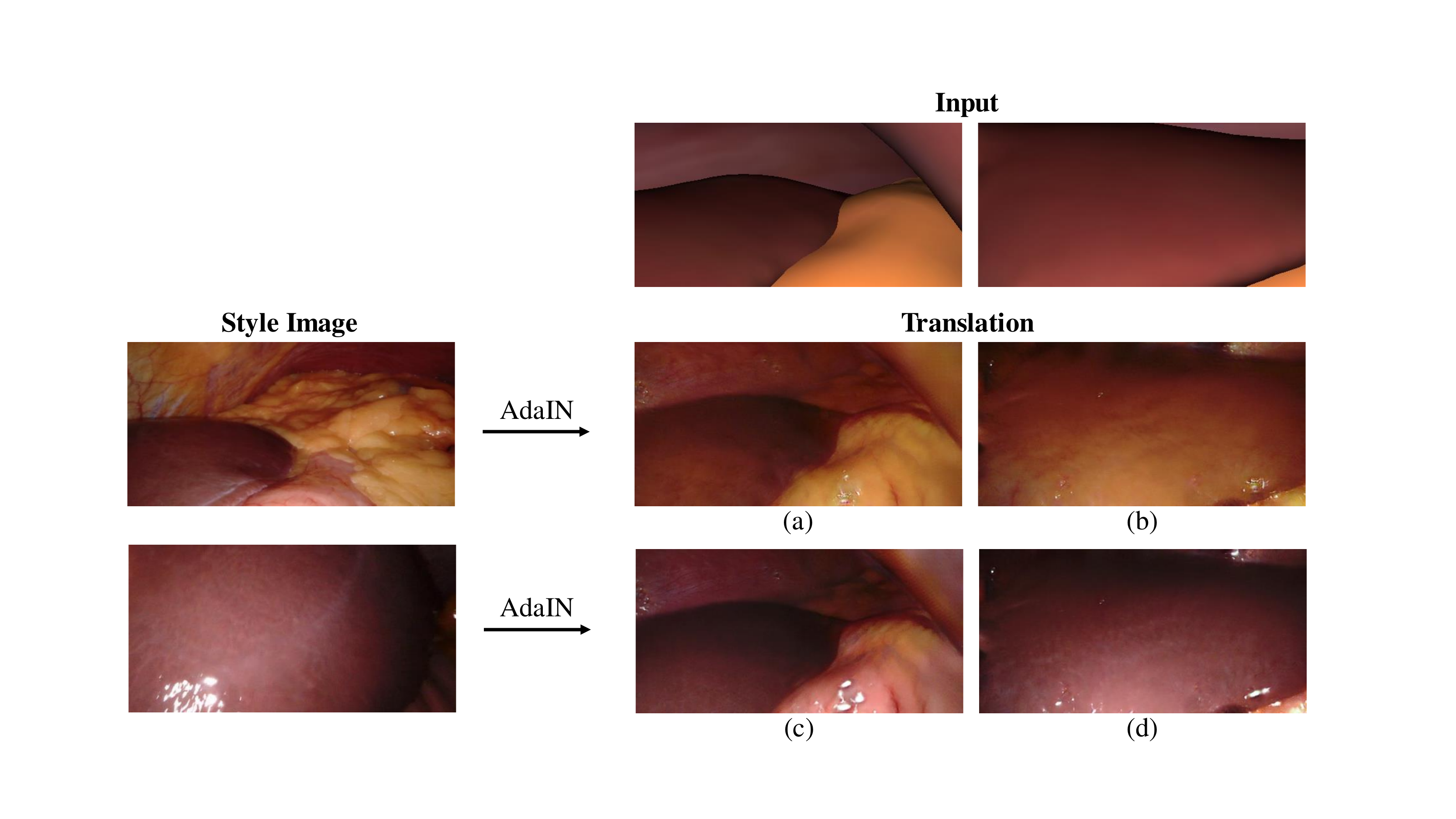}
\end{center}
\caption{In the video setting, \emph{AdaIN} enforces static feature distributions even though the field of view changes over time. We translate two views of the same scene once with a style image of similar content ($a$ and $d$) and once with dissimilar content ($b$ and $c$). We observe that using a style image with a lot of fat forces the model to render yellow color in a view that contains mostly liver ($b$) but produces reasonable results if the content matches ($a)$. We did, however, observe that this effects seems to be most prominent with 'fatty' style images and less drastic in some other cases. Example $c$ shows that the model is able to draw stomach texture on the lower right corner although there seems to be no such texture in the style image. Possibly, the similar hue of liver and stomach allow for this.}
\label{fig:style_in}
\end{figure*}

Secondly, AdaIN styles directly control lower-order statistics (mean, variance) of feature distributions and thereby enforce how much fat, blood, etc. are rendered in a frame\footnote{Fig. 3 of SSIM-MUNIT's supplementary \url{http://opencas.dkfz.de/image2image/supplementary.pdf}}. So, using a single AdaIN style for the whole video enforces a \emph{static} feature distribution (\ie static amount of fat, etc.) across frames although the field of view changes over time. Figure~\ref{fig:style_in} shows how \eg a fatty style image results in translated views with a lot of yellow color regardless of its content. Thus, even if the current view contains only liver, AdaIN styles force the model to render fat. In the image setting, this problem could possibly be circumvented by selecting style images with similar content. In the video setting, however, finding a style that matches all video frames cannot be guaranteed and using multiple styles induces temporal inconsistencies.

For both problems, we believe that introducing styles at texture level would provide a possible solution and that this would be a useful direction for future research.

\end{document}